 \documentclass[pmlr,twocolumn,10pt]{jmlr} 

\usepackage{subcaption}
\usepackage{amsfonts}
\usepackage{amsmath}
\usepackage{makecell}
\newcommand{\ie}{i.e.,\ }

\usepackage{multirow} 
\usepackage{relsize}
\usepackage{threeparttable}
\usepackage{hyperref}

\usepackage{listings}
\usepackage{xcolor}

\lstdefinestyle{mypython}{
  language=Python,
  basicstyle=\ttfamily\footnotesize,
  keywordstyle=\color{blue},
  stringstyle=\color{red!60!black},
  commentstyle=\color{green!50!black},
  showstringspaces=false,
  breaklines=true,
  tabsize=2,
  columns=flexible
}

\usepackage{booktabs}
\usepackage{siunitx}

\usepackage[switch]{lineno}

\theorembodyfont{\upshape}
\theoremheaderfont{\scshape}
\theorempostheader{:}
\theoremsep{\newline}

\jmlrvolume{297}
\jmlryear{2025}
\jmlrworkshop{Machine Learning for Health (ML4H) 2025} 

\title[M\&M-3D: A Data-Efficient Detector for DBT]{From 2D to 3D Without Extra Baggage:\\ Data-Efficient Cancer Detection in Digital Breast Tomosynthesis}

 \author{%
  \Name{Yen Nhi Truong Vu} \Email{nhi@whiterabbit.ai}\\
  \Name{Dan Guo} \Email{dang@whiterabbit.ai}\\
  \Name{Sripad Joshi} \Email{sripad@whiterabbit.ai}\\
  \Name{Harshit Kumar} \Email{harshit@whiterabbit.ai}\\
  \Name{Jason Su} \Email{jason@whiterabbit.ai}\\
  \Name{Thomas Paul Matthews} \Email{tom@whiterabbit.ai}\\
  \addr Whiterabbit.ai, Redwood City, CA, USA
 }

\begin{document}

\maketitle

\begin{abstract}
Digital Breast Tomosynthesis (DBT) enhances finding visibility for breast cancer detection by providing volumetric information that reduces the impact of overlapping tissues; however, limited annotated data has constrained the development of deep learning models for DBT. To address data scarcity, existing methods attempt to reuse 2D full-field digital mammography (FFDM) models by either flattening DBT volumes or processing slices individually, thus discarding volumetric information. Alternatively, 3D reasoning approaches introduce complex architectures that require more DBT training data. Tackling these drawbacks, we propose M\&M-3D, an architecture that enables learnable 3D reasoning while remaining parameter-free relative to its FFDM counterpart, M\&M. M\&M-3D constructs malignancy-guided 3D features, and 3D reasoning is learned through repeatedly mixing these 3D features with slice-level information. This is achieved by modifying operations in M\&M without adding parameters, thus enabling direct weight transfer from FFDM. Extensive experiments show that M\&M-3D surpasses 2D projection and 3D slice-based methods by 11--54\% for localization and 3--10\% for classification. Additionally, M\&M-3D outperforms complex 3D reasoning variants by 20--47\% for localization and 2--10\% for classification in the low-data regime, while matching their performance in high-data regime. On the popular BCS-DBT benchmark, M\&M-3D outperforms previous top baseline by 4\% for classification and 10\% for localization.




\end{abstract}

\paragraph*{Data and Code Availability}
 We utilize a multi-site in-house DBT dataset. This dataset was collected from 14 U.S. sites. We also evaluate on the publicly available BCS-DBT test set \citep{buda2021data}. Code is provided in Appendix \ref{apd:code}.

\paragraph*{Institutional Review Board (IRB)} The research does not require IRB approval.

\section{Introduction}
Breast cancer is a leading cause of cancer-related mortality among women \citep{ferlay2020global}, with early detection critical for patient outcomes \citep{duffy2020mammography}. Digital Breast Tomosynthesis (DBT) is increasingly adopted as an alternative to Full-Field Digital Mammography (FFDM) due to its offering of volumetric data that reduces the impact of overlapping tissues and enhances finding visibility \citep{friedewald2014tomo,richman2021comparative}. Yet, while deep learning has shown strong performance in FFDM, progress in DBT has been more limited due to two main reasons: (1) since DBT is a relatively recent technology, data remain scarce, and (2) the high cost of annotating DBT volumes further hinders the training of robust deep learning models.

\begin{figure*}[t]
    \centering
    \includegraphics[width=\textwidth]{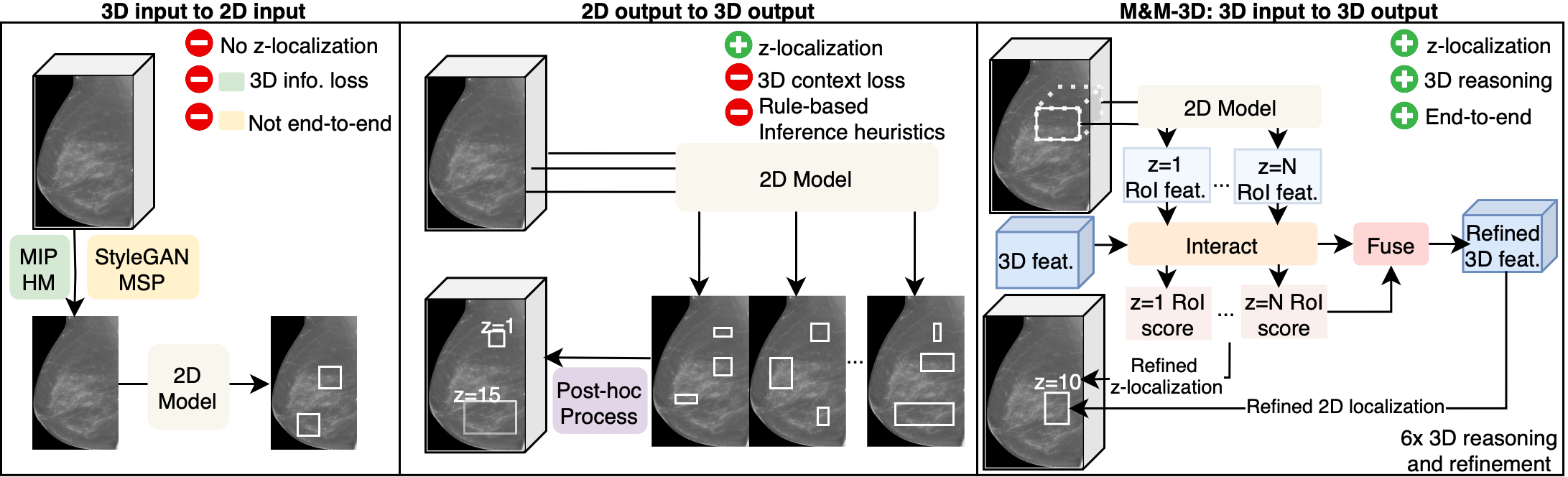}
    \caption{Common DBT approaches either (left) project the input into 2D, making z-localization impossible, or (middle) process the volume slice by slice, relying on heuristics for output aggregation. M\&M-3D (right) enables seamless 3D reasoning by dynamically fusing slice-level features into 3D representations, which repeatedly interact with the slices to facilitate 3D information mixing.}
    \label{fig:intro}
\end{figure*}

A key strategy to address data limitations is to maximize the use of FFDM-pretrained models. For example, prior work reduces the 3D input to 2D (\autoref{fig:intro}, left), such as via maximum intensity projection (MIP) \citep{samala2014digital}, histogram matching (HM) \citep{singh2020adaptation}, StyleGAN \citep{jiang2019synthesize,jiang2021synthesis} and maximum suspicious projection (MSP) \citep{lotter2021robust}. 
This strategy makes $z$-axis localization impossible and risks 3D information loss. The most widely used alternative is to train with one slice at a time and then convert slice-by-slice 2D outputs into 3D outputs during inference (\autoref{fig:intro}, middle) \citep{Alberb2024CoMoTo,buda2021data,fan2019computer,konz2023competition,lai2020dbt,li2021deep}. This approach struggles to leverage the full 3D context and increases the number of candidate boxes that persist as false positives despite using post-hoc aggregation, such as Non-Maximum Suppression (NMS) \citep{ge2024non,luo2021nms,yu2023multiple}. Conversely, recent methods introduce complex reasoning modules to incorporate 3D information \citep{lee2023transformer,park2023efficient,tardy2021trainable,zhang2018classification}. However, they come with more parameters not pretrained on FFDM and thus may struggle in the low-data regime.


Our work aims to tackle these limitations by designing an architecture that can perform \emph{learnable} 3D reasoning while remaining \emph{parameter-free} relative to its 2D counterpart, \ie  without introducing new modules that prevent direct transfer from FFDM and increase data requirements. To this end, we propose M\&M-3D (\autoref{fig:intro}, right), a simple yet effective extension of M\&M \citep{truong2023m}, a high-performing 2D mammography detector. M\&M-3D fuses slice-level features into 3D features using a malignancy-driven weighting mechanism. These 3D features repeatedly interact with individual slices in the model's cascading heads using existing dynamic convolution modules. This design maintains volumetric awareness, provides implicit $z$-axis localization, and crucially enables 3D reasoning without adding parameters beyond the original 2D architecture. Through extensive experiments, we demonstrate two key findings:


\paragraph{1. M\&M-3D provides learnable 3D reasoning without additional parameters.} By preserving 3D information, M\&M-3D outperforms 2D projection methods by 0.27 in recall at 0.25 false positives per volume (R@0.25) and 0.08 in area under the receiver operating characteristic curve (AUC). By learning 3D reasoning instead of relying on aggregation heuristics, M\&M-3D outperforms slice-by-slice baselines by 0.09 in R@0.25 and 0.03 in AUC.
\paragraph{2. M\&M-3D provides efficient 3D reasoning in low data regime.} By avoiding additional 3D-specific parameters, M\&M-3D maximizes the use of FFDM-pretrained weights. M\&M-3D matches performance of complex 3D variants at 100\% data and outperforms them by 0.10--0.20 in R@0.25 and 0.02--0.09 in AUC at 10\% data.

Finally, we also evaluate M\&M-3D on the widely used BCS-DBT benchmark \citep{buda2021data}. M\&M-3D demonstrates excellent generalizability, achieving 0.97 AUC and 0.83 R@0.25, the highest results reported to date (Appendix~\ref{apd:duke}). 


\begin{figure*}[t]
\includegraphics[width=\textwidth]{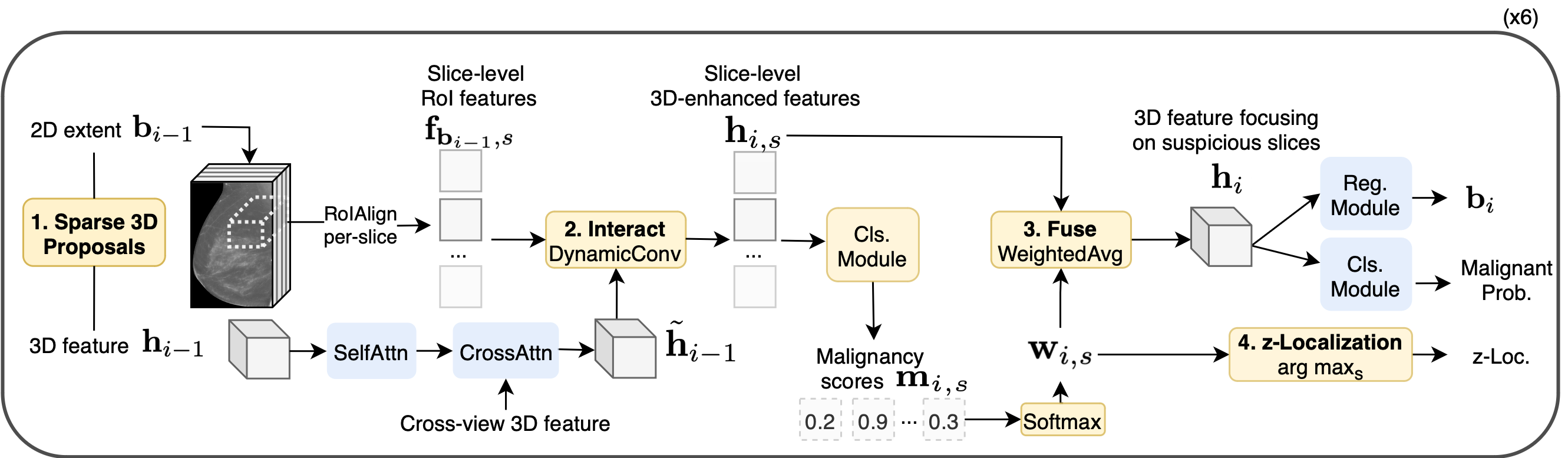}
\caption{M\&M-3D extends M\&M (blue, \autoref{sec:prereq}) with parameter-free 3D reasoning (yellow, \autoref{sec:methods}). 
3D proposals, parameterized by 3D features $\mathbf{h}_{i-1}$ and 2D extent $\mathbf{b}_{i-1}$ spanning all slices $s$, are refined using 6 cascade heads ($1 \leq i \leq 6$). These 3D features interact with 2D RoI features $\mathbf{f}_{\mathbf{b}_{i-1},s}$ 
across slices, producing slice-level object features $\mathbf{h}_{i,s}$ enhanced with 3D context. 
The classification (Cls.) module is reused to produce finding-slice scores $\mathbf{w}_{i,s}$, which are used to 
fuse $\mathbf{h}_{i,s}$ into refined 3D features $\mathbf{h}_{i}$ focusing on the most suspicious slices. 
$z$-axis localization is obtained as the slice with maximum score, i.e., $\arg\max_{s} \mathbf{w}_{i,s}$.}
\label{fig:arch}
\end{figure*}

\section{Prerequisites} \label{sec:prereq}
\paragraph{Sparse R-CNN.}  \cite{sun2021sparse} proposed Sparse-RCNN, which utilizes a sparse set of $N$ proposals parametrized by boxes $\mathbf{b}_0 \in \mathbb{R}^{N \times 4}$ and corresponding features $\mathbf{h}_0 \in \mathbb{R}^{N \times D}$, with $D$ the hidden dimension.
These proposals are refined through 6 cascading heads, where the $i^\text{th}$ head refines $\mathbf{h}_{i-1}$ into new features $\mathbf{h}_i$ using two key operations
\begin{align}
&\mathbf{h}'_{i-1} = \mathrm{SelfAttn}(\mathbf{h}_{i-1}), \label{eq:selfattn}\\
&\mathbf{h}_i = \mathrm{DynamicConv}(\mathbf{h}'_{i-1}, \mathbf{f}_{\mathbf{b}_{i-1}}), \label{eq:dynconv}
\end{align}
where $\mathbf{f}_{\mathbf{b}_{i-1}} \in \mathbb{R}^{N \times k^2 \times D}$ are region-of-interest (RoI) features pooled from boxes $\mathbf{b}_{i-1}$ and $k$ is the pool size. Self-attention ($\mathrm{SelfAttn}$) enables global feature mixing across proposals. Dynamic convolutions ($\mathrm{DynamicConv}$) update $\mathbf{f}_{\mathbf{b}_{i-1}}$ using kernels dynamically generated by $\mathbf{h}'_{i-1}$, thus conditioning each proposal on its corresponding RoI features. A regression module is applied to $\mathbf{h}_i$ to obtain new boxes $\mathbf{b}_i$, completing the iterative refinement.

\paragraph{M\&M.} \cite{truong2023m} proposed M\&M, which adapts Sparse R-CNN to 2D mammography, where each breast is imaged under two complementary views: cranio-caudal (CC) and medio-lateral oblique (MLO). Let $\mathrm{CrossAttn}(\mathbf{q},\mathbf{k})$ denote cross-attention from query $\mathbf{q}$ to key/value $\mathbf{k}$. Let $\mathbf{h}'^{a}_{i-1}$ denote the alternative view's post-SelfAttn feature. M\&M introduces multi-view attention to mix information from the alternative view as follows:
\begin{align}
\tilde{\mathbf{h}}_{i-1} &= \mathrm{CrossAttn}(\mathbf{h'}_{i-1}, \mathbf{h}'^{a}_{i-1}), \label{eq:crossattn} \\
\mathbf{h}_i &= \mathrm{DynamicConv}(\tilde{\mathbf{h}}_{i-1}, \mathbf{f}_{\mathbf{b}_{i-1}}). \label{eq:dynconvmm}
\end{align}
M\&M also employs a multi-instance learning (MIL) formulation. Proposal malignancy scores $\mathbf{m}_i$ are aggregated by noisy-or pooling to yield image scores, and mean pooling to yield breast scores. Finding-level losses are disabled if annotations are absent.

\section{M\&M-3D Architecture}
\label{sec:methods}

Our goal is to achieve learnable 3D reasoning, but without adding any additional learnable parameters. This design decision will allow us to maximize the pre-trained weights from the FFDM model. To implement such a design, we propose the following parameter free extensions of M\&M (\autoref{fig:arch}):


\paragraph{1. Sparse 3D proposals.}  Consider a DBT volume with $S$ slices. A naive slice-by-slice application of M\&M would yield an unstructured set of $N \times S$ 2D proposals. These 2D proposals are independent across slices, which complicates the learning process and necessitates post-hoc aggregation to consolidate 2D detections into 3D objects (\autoref{fig:intro}, middle). To enforce spatial coherence, we reinterpret the initial proposals as \emph{3D primitives}: in M\&M-3D, $\mathbf{h}_0 \in \mathbb{R}^{N \times D}$ represents features associated with 3D proposals with 2D extents $\mathbf{b}_0 \in \mathbb{R}^{N \times 4}$ and depths that span the entirety of the DBT volume. We retain 6 cascading heads, where the $i^\text{th}$ head incorporates slice features to refine the previous 3D feature $\mathbf{h}_{i-1}$ and box $\mathbf{b}_{i-1}$ into better 3D feature $\mathbf{h}_i$ and box $\mathbf{b}_i$. With this new interpretation, M\&M-3D preserves a sparse set of $N$ 3D proposals throughout refinement.



\paragraph{2. Slice-level feature interaction.} During refinement, we reuse SelfAttn (\autoref{eq:selfattn}) and CrossAttn (\autoref{eq:crossattn}) to obtain $\tilde{\mathbf{h}}_{i-1}$, a 3D proposal feature enhanced with information from other proposals across both views. 
DynamicConv (\autoref{eq:dynconv}) now convolves slice-level features using kernels dynamically generated by 3D feature $\tilde{\mathbf{h}}_{i-1}$, effectively allowing for 2D and 3D feature interaction.

Concretely, in the 2D setting, DynamicConv (\autoref{eq:dynconv}) operates between proposal features $\tilde{\mathbf{h}}_{i-1}$ and RoI features $\mathbf{f}_{\mathbf{b}_{i-1}} \in \mathbb{R}^{N\times k^2 \times D}$. For DBT, we apply RoIAlign across all slices $1 \leq s \leq S$, obtaining slice-specific features $\mathbf{f}_{\mathbf{b}_{i-1},s}$. We then reuse the same dynamic convolution module as follows
\begin{equation}
    \mathbf{h}_{i,s} = \text{DynamicConv}(\tilde{\mathbf{h}}_{i-1}, \mathbf{f}_{\mathbf{b}_{i-1},s}).
\end{equation}
Here, $\mathbf{h}_{i,s} \in \mathbb{R}^{N\times D}$ represents 3D-enhanced slice-level features: while remaining specific to slice $s$, $\mathbf{h}_{i,s}$ now contains 3D context carried over from $\tilde{\mathbf{h}}_{i-1}$. This design is both simple and effective---by reusing dynamic convolutions, M\&M-3D seamlessly adapts to 3D reasoning without adding parameters. More importantly, since this interaction occurs in every head, proposals repeatedly mix volumetric information, ensuring progressive refinement of 3D spatial cues.


\paragraph{3. Slice-to-volume feature fusion.} The vector $[\mathbf{h}_{i,1}, \cdots, \mathbf{h}_{i,S}]$ contains enhanced features of the region $\mathbf{b}_{i-1}$ across all slices; thus, it is a natural candidate from which to derive $\mathbf{h}_i \in \mathbb{R}^{N\times D}$, the refined 3D representation. To this end, we propose a simple weighted average (Wgt. Avg.) fusion module. First, we pass $\mathbf{h}_{i,s}$ through the classification module of M\&M to obtain malignancy scores of each finding across slices $\mathbf{m}_{i,s}.$ We then compute
\begin{equation} \mathbf{h}_i = \sum_{s=1}^S \mathbf{w}_{i,s} \odot \mathbf{h}_{i,s}, \qquad \mathbf{w}_{i,s} = \frac{e^{\mathbf{m}_{i,s}}}{\sum_{s'=1}^S e^{\mathbf{m}_{i,s'}}} \in \mathbb{R}^{N}, \end{equation}
where $\odot$ denotes element-wise multiplication. The refined vector $\mathbf{h}_{i}$ is a genuine 3D proposal feature: it contains information from all slices, where slices with higher suspicion contribute more strongly to the representation through their larger weights. Furthermore, for different proposal $n \in \{1, \dots, N\}$, the weights across slices $\mathbf{w}_{i,\cdot}[n]$ are different, allowing each proposal to specialize in the subset of slices most relevant to its RoI.

\paragraph{4. $z$-axis localization.} In clinical practice, radiologists primarily make callback and diagnostic decisions based on the slice where the finding is most clearly visible (see Appendix \ref{apd:usability}). Following this workflow, our design focuses supervision on the most suspicious slice rather than the visibility range of findings. Concretely, the scores $\{\mathbf{w}_{i,s}\}_{s=1}^S$ encode finding malignancy across slices, implicitly providing z-axis localization. In particular, $\mathbf{z}_i = \arg\max_{s} \mathbf{w}_{i,s} \in \mathbb{R}^N$ is taken as the most suspicious slice for the findings. Given a proposal $m$ matching a ground truth finding, if annotation of the most suspicious slice $z$ is available, we can apply the cross entropy loss $-\log(\mathbf{w}_{i,z}[m])$. We do not apply losses on the remaining slices $\mathbf{w}_{i,s}$ with $s \neq z$ since it is not known whether the finding is visible in such slice. Thus, the total training loss for M\&M-3D is
\begin{equation}
    \mathcal{L} = \mathcal{L}_{\text{M\&M}} - \mathbf{1}_{\text{annotated finding}} \sum_{i=1}^6 \log(\mathbf{w}_{i,z}[m]),
\end{equation}
where $\mathcal{L}_{\text{M\&M}}$ contains finding-level 2D localization loss and image and breast-level classification loss.

\begin{figure*}[t]
\centering\includegraphics[width=\textwidth]{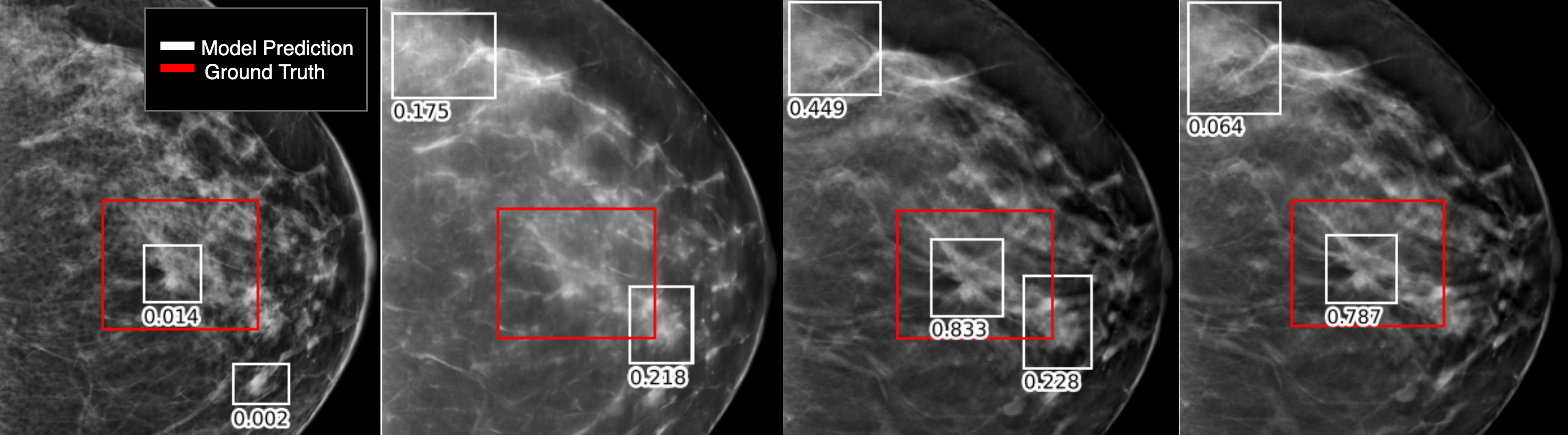}
    \caption{Qualitative results. From left to right: (1) \textbf{FFDM}: The finding is obscured by surrounding tissue, leading to low detection scores. (2) \textbf{MIP} still suffers from occlusion and introduces false positives. (3) \textbf{Buda*} assigns a high score to the correct finding on the most suspicious slice but also increases the number of false positives. (4) \textbf{M\&M-3D} successfully assigns a high score to the malignant finding on the most suspicious slice while maintaining low scores in other regions.}
    \label{fig:qualitative}
\end{figure*}
\begin{table}
    \footnotesize
    \setlength{\tabcolsep}{3pt}
    \centering
    \caption{The number of malignant, benign and negative breasts in each dataset split. Last column reports the number of breasts with finding annotations.}
    \label{tab:data_distribution}
    \begin{tabular}{lcccc}
        \toprule
        \textbf{Split} & \textbf{Malignant} & \textbf{Benign} & \textbf{Negative} & \textbf{Annotated} \\
        \midrule
        Train & 2311 & 8524 & 10266 & 955 \\
        Val   & 302  & 1100 & 1489  & 246 \\
        Test  & 664  & 3462 & 2677  & 204 \\
        \bottomrule
    \end{tabular}
\end{table}

\section{Experiments}
M\&M-3D avoids new parameters to fully leverage FFDM-pretraining. We compare against parameter-free baselines (\autoref{sec:quantitative_performance}), evaluate data efficiency, including against parameter-heavy variants (\autoref{sec:data_efficiency}), and present ablation studies (\autoref{sec:ablation_studies}).
\subsection{Experiment details}
\label{sec:experiment_details}
\paragraph{Implementation details.} To reduce memory, we crop the background region and resize the image to a width of 1100 and max length of 2200. We also downsample the z-resolution to 16 using MaxPool, corresponding to a resizing factor of 1.25–8.50. During inference, the z-coordinate of each prediction is mapped back by selecting the middle slice of the corresponding downsampled region. Horizontal and vertical flip augmentations are applied during training. We use AdamW optimizer with $2.5\times 10^{-5}$ learning rate and 0.0001 weight decay. Unless stated otherwise, all models are initialized with FFDM pretrained weights shared by authors of M\&M.

\paragraph{Dataset.} We utilize a multi-site in-house DBT dataset. This dataset was collected from 14 sites and was split into training, validation, and test sets at the site level. \autoref{tab:data_distribution} summarizes dataset statistics for each split. Findings are labeled with bounding boxes on the most visible slice, and for the test split, their visibility range across slices is also annotated.

\paragraph{Metric.} We report recall at $X$ false postive per volume (R@$X$) with $X \in \{0.25, 0.5\}$. A proposal is a true positive if the intersection over union (IoU) of the 2D proposal and 2D ground truth is at least 0.25 and the predicted slice is within the visibility range. For classification, we report area under the receiver operating characteristic curve (AUC). Standard errors are bootstrapped for R@0.25 and R@0.5, and computed via \cite{delong1988comparing} for AUC.


\subsection{M\&M-3D provides 3D reasoning without additional parameters}
\label{sec:quantitative_performance}
An important design motivation for M\&M-3D is to maximize transfering ability from FFDM to DBT, thus we do not introduce any additional parameters. This allows the model to operate even without training on DBT data. In this section, we evaluate M\&M-3D against other parameter-free methods and show that it provides meaningful 3D reasoning where prior approaches fall short.

\paragraph{Parameter-free baselines.}
We compare against three representative baselines:
\subparagraph{1. 2D projection methods.} Maximum Intensity Projection (MIP)~\citep{samala2014digital} and Histogram Matching (HM)~\citep{singh2020adaptation} collapse the DBT volume into a single FFDM-like image.
\subparagraph{2. Slice-based aggregation methods.} The most common DBT strategy trains slice-level models, then aggregates outputs during inference  \citep{Alberb2024CoMoTo,buda2021data,fan2019computer,konz2023competition,lai2020dbt,li2021deep}. This approach relies on sampling positive slices during training. Since 60\% of our malignant volumes are unannotated (\autoref{tab:data_distribution}), we assess two versions of the representative baseline by \cite{buda2021data}: (1) \textit{Buda}, which samples a random slice when annotations are unavailable, risking noisy supervision, and (2) \textit{Buda*}, which discards all unannotated malignant volumes, drastically reducing training data.
\subparagraph{3. FFDM baseline.} Because our dataset provides paired DBT and FFDM images, we also report M\&M performance when evaluated directly on FFDM. This acts as a non-transfer benchmark, reflecting performance without cross-modality adaptation, and serves as reference for evaluating transfer to DBT.

\paragraph{Inference-only performance (\autoref{tab:baseline_inference}).}
Projection baselines perform poorly because collapsing volumes into 2D images discards depth information and amplifies occlusion (\autoref{fig:qualitative}, second column). By preserving depth information, M\&M-3D improves over MIP/HM by $0.20$–$0.39$ R@0.25 and $0.08$–$0.20$ AUC. For 3D localization, M\&M-3D performs on par with Buda (R@0.25: $0.46$ vs.\ $0.45$, AUC: $0.85$ vs.\ $0.84$), indicating that even without finetuning, M\&M-3D's built-in 3D reasoning capability is already as effective as post-hoc aggregation methods (\autoref{tab:baseline_inference}). While the FFDM baseline remains the strongest in this regime, M\&M-3D substantially closes the gap, demonstrating its ability to transfer from FFDM to DBT.

\begin{table}
\footnotesize
\centering
\setlength{\tabcolsep}{3pt}
\caption{Inference-only performance.}
\label{tab:baseline_inference}
\begin{tabular}{llccc}
\toprule
\textbf{Method} & \textbf{Input} & \textbf{R@0.25} & \textbf{R@0.5} & \textbf{AUC} \\
\midrule
\multicolumn{5}{l}{\emph{2D localization (2D $\text{IoU}\geq 0.25$)}} \\
M\&M & FFDM & \textbf{0.61$_{\pm0.025}$} & \textbf{0.67$_{\pm0.025}$} & \textbf{0.89$_{\pm0.009}$} \\
MIP & Proj. & 0.14$_{\pm0.018}$ & 0.16$_{\pm0.019}$ & 0.65$_{\pm0.012}$ \\
HM & Proj. & 0.33$_{\pm0.024}$ & 0.41$_{\pm0.026}$ & 0.77$_{\pm0.010}$ \\
\textbf{M\&M-3D} & Vol. & 0.53$_{\pm0.025}$ & 0.57$_{\pm0.025}$ & 0.85$_{\pm0.008}$ \\
\midrule
\multicolumn{5}{l}{\emph{3D localization (2D IoU $\geq 0.25$ + slice in visible range)}} \\
Buda  & Vol. & 0.45$_{\pm0.026}$ & \textbf{0.49$_{\pm0.026}$} & 0.84$_{\pm0.009}$ \\
\textbf{M\&M-3D} & Vol. & \textbf{0.46$_{\pm0.026}$} & 0.48$_{\pm0.026}$ & \textbf{0.85$_{\pm0.008}$} \\
\bottomrule
\end{tabular}
\end{table}

\begin{table}
\footnotesize
\centering
\setlength{\tabcolsep}{3pt}
\caption{Fine-tuned performance. (*) models are trained without unannotated malignant volumes.}
\label{tab:baseline_finetuned}
\begin{tabular}{llccc}
\toprule
\textbf{Method} & \textbf{Input} & \textbf{R@0.25} & \textbf{R@0.5} & \textbf{AUC} \\
\midrule
\multicolumn{5}{l}{\emph{2D localization (2D $\text{IoU}\geq 0.25$)}} \\
M\&M & FFDM & 0.62$_{\pm0.027}$ & 0.69$_{\pm0.026}$ & 0.89$_{\pm0.008}$ \\
MIP & Proj. & 0.52$_{\pm0.026}$ & 0.60$_{\pm0.025}$ & 0.86$_{\pm0.008}$ \\
HM & Proj. & 0.53$_{\pm0.027}$ & 0.60$_{\pm0.026}$ & 0.87$_{\pm0.008}$ \\
\textbf{M\&M-3D} & Vol. & \textbf{0.80$_{\pm0.020}$} & \textbf{0.84$_{\pm0.019}$} & \textbf{0.95$_{\pm0.004}$} \\
\midrule
\multicolumn{5}{l}{\emph{3D localization (2D IoU $\geq 0.25$ + slice in visible range)}} \\
Buda* & Vol. & 0.68$_{\pm0.025}$ & 0.74$_{\pm0.023}$ & 0.92$_{\pm0.006}$ \\
Buda & Vol. & 0.68$_{\pm0.023}$ & 0.73$_{\pm0.023}$ & 0.92$_{\pm0.006}$ \\
M\&M-3D* & Vol. & 0.73$_{\pm0.023}$ & 0.79$_{\pm0.021}$ & 0.94$_{\pm0.005}$ \\
\textbf{M\&M-3D} & Vol. & \textbf{0.77$_{\pm0.021}$} & \textbf{0.82$_{\pm0.019}$} & \textbf{0.95$_{\pm0.004}$} \\
\bottomrule
\end{tabular}
\end{table}

\paragraph{Fine-tuning performance (\autoref{tab:baseline_finetuned}).}
With supervision, the advantages of explicit 3D reasoning become more pronounced. M\&M-3D surpasses the FFDM model by $0.18$ R@0.25 and $0.06$ AUC, reflecting its ability to exploit lesions that are visible in DBT slices but obscured in FFDM projections (\autoref{fig:qualitative}, left). It also outperforms projection baselines by ${\sim}0.25$ in recall and ${\sim}0.09$ in AUC, demonstrating that simple 2D proxies cannot capture volumetric detail. Against slice-based methods, M\&M-3D outperforms both Buda variants by about 0.10 in R@0.25, R@0.5 and by 0.03 in AUC. Even when evaluated under the same data constraints, M\&M-3D* remains superior to Buda*, highlighting the advantage of learned 3D reasoning over heuristic aggregation. Notably, Buda gains no benefit over Buda* from including unannotated malignant volumes due to the noise introduced by random slice sampling. Meanwhile, M\&M-3D significantly outperforms M\&M-3D*, proving its effective use of unannotated data. The key advantage of M\&M-3D is evident in \autoref{fig:qualitative}: while slice-based models introduce more false positives during aggregation, M\&M-3D leverages learned 3D reasoning to effectively filter out spurious detections.

\begin{figure}[t]
    \centering
    \includegraphics[width=0.95\linewidth]{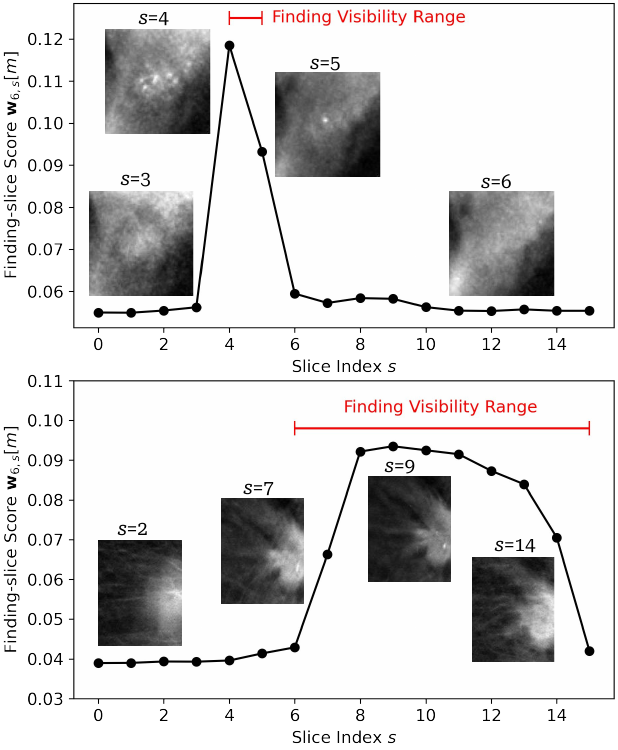}
    \caption{Finding-slice scores produced by M\&M-3D. For a malignant finding, we identify the highest scoring proposal $m$ that matches it, and plot the scores $\mathbf{w}_{6,s}[m]$. The red bar denotes the range of slices where the finding is visible based on ground truth annotations. Top: a small cluster of calcifications visible in only 2/16 slices, yielding a sharp localized peak in $\mathbf{w}_{6,s}[m]$. Bottom: a mass visible on 10/16 slices, yielding a broad span of elevated $\mathbf{w}_{6,s}[m]$ values. Insets show representative slices.}
    \label{fig:zaxis_qualitative}
\end{figure}

\paragraph{Interpretable $z$-axis localization.} \autoref{fig:zaxis_qualitative} illustrates that the finding-slice scores $\mathbf{w}_{i,s}$ provide an interpretable signal of how the model localizes findings along the $z$-axis. In particular, the scores align closely with finding's visibility in DBT slices. In the top example, a small cluster of calcifications is visible in only 2 of 16 slices, and the score distribution exhibits a sharp, localized peak precisely within this range. In the bottom example, a mass extends across 10 of 16 slices, and the score profile remains elevated across the full span before tapering off as the lesion disappears. These patterns demonstrate that M\&M-3D adapts its scoring to the lesion’s extent rather than relying on fixed aggregation heuristics.

\paragraph{Generalizability.} To assess generalizability, we evaluate  M\&M-3D on the BCS-DBT test set \citep{buda2021data}. Without training on DBT, M\&M-3D achieves 0.91 AUC and 0.56 R@0.5, comparable to top performing methods \citep{konz2023competition,terrassin2024thick}. After finetuning, M\&M-3D achieves 0.97 AUC, 0.83 R@0.25 and 0.85 R@0.5. To our knowledge, these are the highest performance results on this dataset (Appendix \ref{apd:duke}).

\subsection{M\&M-3D provides efficient 3D reasoning in low data regime} 
\label{sec:data_efficiency}
Given that DBT model development is hindered by scarce data and costly annotation, we assess the data efficiency of M\&M-3D in two ways: (1) by reducing the fraction of training data to simulate limited data availability, and (2) by reducing the proportion of annotated data while keeping the dataset size fixed at 100\% to simulate limited annotation availability. When varying data size, we report both detection (R@0.25) and classification metrics (AUC). When varying annotation size, we discard only box annotations while keeping all images available. This keeps classification performance (AUC) stable across models, so we report only detection performance.

\paragraph{Complex 3D reasoning baselines.} We benchmark M\&M-3D against alternative designs that incorporate more sophisticated trainable components for 3D reasoning. These baselines have the potential to capture richer 3D representations, but at the cost of being more data-hungry.

\subparagraph{1. TimeSform.} 
Following \cite{lee2023transformer}, \textit{TimeSform} performs joint spatial and depth attention on slice-level RoI features $\mathbf{f}_{\mathbf{b}_{i-1},s}$ \citep{bertasius2021space}. 
Attention is applied to fuse RoI features across slices: 
\begin{equation}
\mathbf{f}_{\mathbf{b}_{i-1}} = \text{Attn}\!\left(\{\mathbf{f}_{\mathbf{b}_{i-1},s}\}_{s=1}^S\right), \label{eq:fpnattn}
\end{equation}
where $\mathbf{f}_{\mathbf{b}_{i-1}} \in \mathbb{R}^{N \times k^2 \times D}$ encodes volumetric RoI context. From here, $\mathrm{DynamicConv}$ can be applied directly as in \autoref{eq:dynconvmm}. $z$-axis localization is derived from the attention weights of \autoref{eq:fpnattn}.

\subparagraph{2. QuerySummary} utilizes learnable queries that act as adaptive pooling operators \citep{devlin2019bert} to perform slice-to-volume feature fusion: 
\begin{equation}
\mathbf{h}_i = \text{Attn}\!\left(\mathbf{q}_i, \{\mathbf{h}_{i,s}\}_{s=1}^S \right), \label{eq:querysum}
\end{equation}
where $\mathbf{q}_i \in \mathbb{R}^{N \times D}$ are learnable query embeddings. Similar to M\&M-3D, the attention weights from \autoref{eq:querysum} are reused as finding–slice scores $\mathbf{w}_i$.

\subparagraph{3. MLP-Regress} replaces our malignancy-guided weighted average with an MLP-based summarization:
\begin{equation}
\mathbf{h}_i = \text{MLP}\!\left(\{\mathbf{h}_{i,s}\}_{s=1}^S\right),
\end{equation}
and extends the regression module to explicitly predict central slices $\mathbf{z}_i = \text{Regress}\!\left(\mathbf{h}_i\right)$.

\begin{figure}[t]
    \centering
    \includegraphics[width=0.85\linewidth]{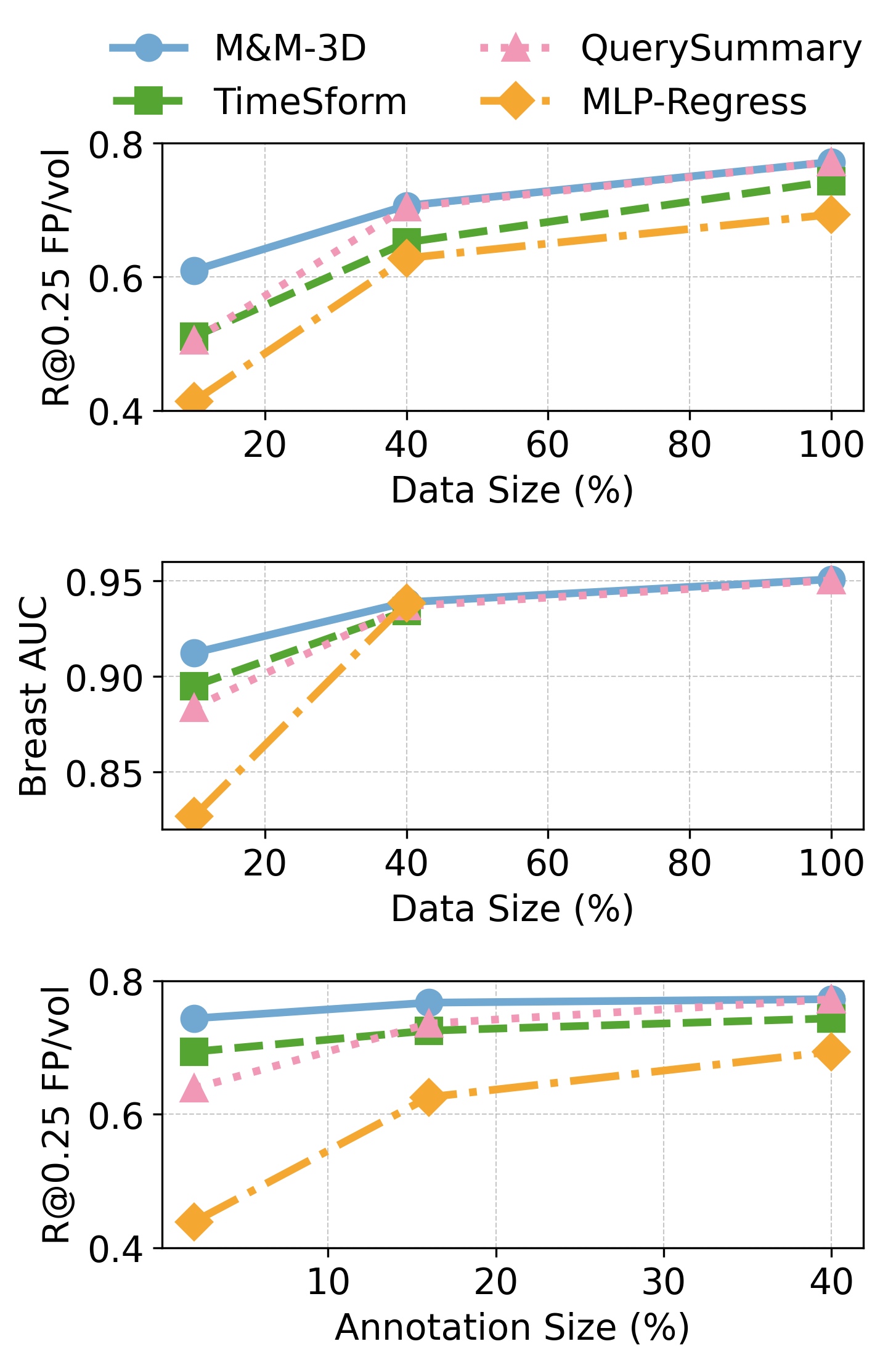}
    \caption{Comparison of M\&M-3D with complex 3D reasoning variants. M\&M-3D performs similarly to all alternatives in high-data regime but outperforms them significantly in low-data regime, illustrating its data efficiency. See Appendix \ref{apd:fig5} for figure data.}
    \label{fig:3d_comparison}
\end{figure}
\paragraph{Performance under limited data.}
As shown in \autoref{fig:3d_comparison}, M\&M-3D consistently outperforms all complex variants in low-data regimes. With only 10\% of training data ($\sim$230 malignant breasts), M\&M-3D achieves at least 0.10 gain in R@0.25 and 0.02 in AUC compared to all alternatives. Notably, in this extreme setting, M\&M-3D is the only method to surpass the FFDM-only M\&M baseline (\autoref{tab:baseline_finetuned}, row 1). This highlights M\&M-3D's superior data efficiency, making it an ideal DBT solution for institutions with large FFDM datasets but limited DBT data.

\begin{figure}[t]
    \centering
    \includegraphics[width=0.85\linewidth]{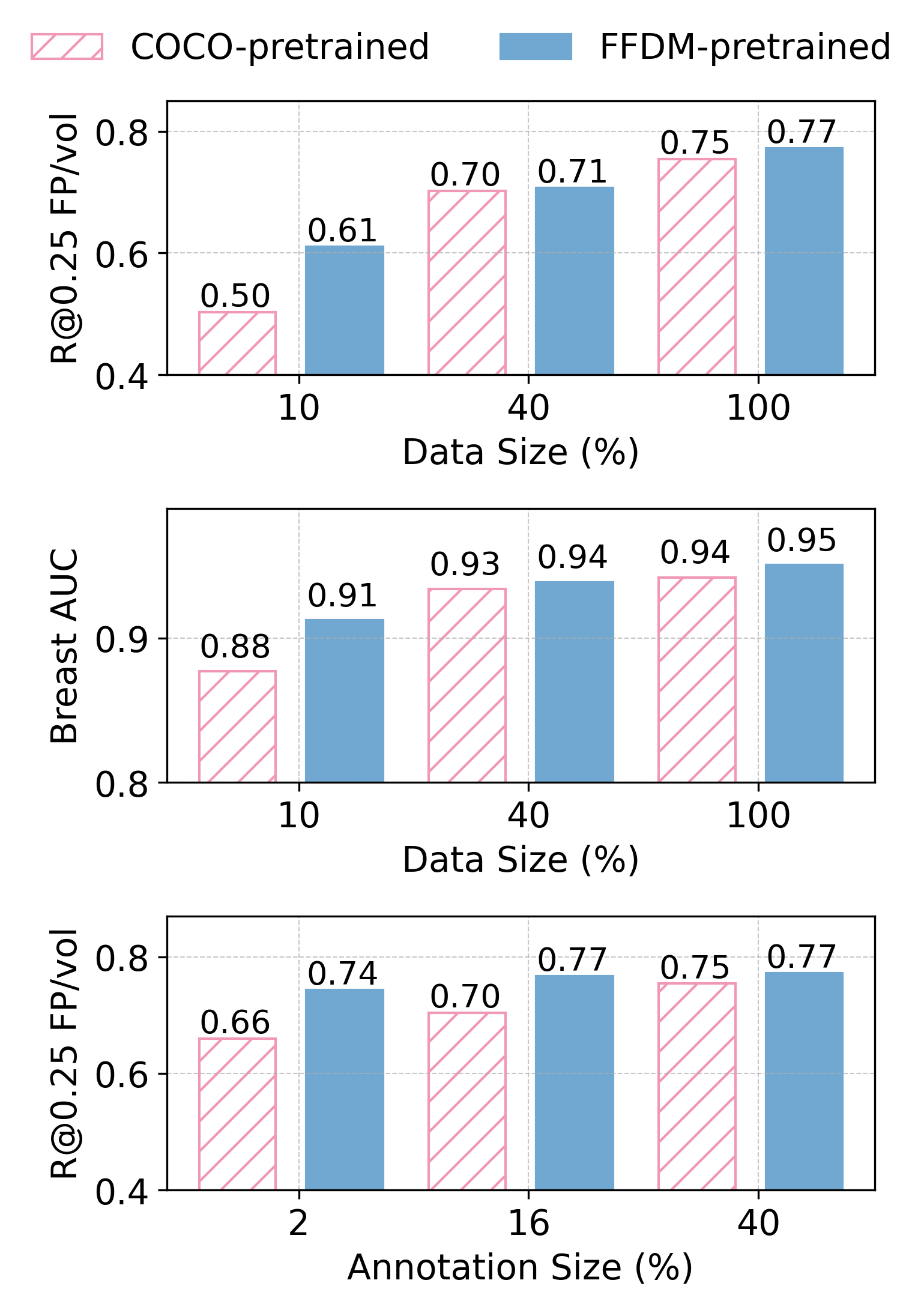}  
    \caption{M\&M-3D with FFDM vs. COCO pretrained weights. FFDM-pretrained models outperform COCO-pretrained models in detection and classification across all dataset and annotation sizes, with the largest gaps in low-data regimes.}
    \label{fig:coco_comparison}
\end{figure}
\paragraph{Performance under limited annotations.} \autoref{fig:3d_comparison}, last row illustrates detection performance when varying the fraction of annotated cases while fixing the dataset size. At maximum, 40\% of malignant cases in our training set is annotated (\autoref{tab:data_distribution}), and we further simulate settings with only 2\% and 16\% annotated cases. Again, we observe larger performance gap when annotations are more limited: at 2\% annotations ($\sim$50 breasts), M\&M-3D achieves 0.74 R@0.25, while the strongest baseline trails at 0.70. M\&M-3D's ability to succeed with minimal annotations is especially beneficial for institutions that may already possess DBT data but lack the resources or expertise to annotate them extensively.


%


%

\paragraph{Benefits of transfer learning.} We examine the impact of initializing M\&M-3D with FFDM-pretrained weights versus COCO-pretrained weights \citep{lin2014microsoft}. \autoref{fig:coco_comparison} shows FFDM pretraining consistently outperforms COCO pretraining, achieving higher detection and classification performance across all data sizes. The gap is largest at 10\% data ($\sim$230 malignant breasts), where FFDM pretraining improves R@0.25 by 0.11 and AUC by 0.03. FFDM pretraining also boosts annotation efficiency (\autoref{fig:coco_comparison}, bottom): at 2\% annotations ($\sim$50 breasts), M\&M-3D initialized with FFDM weights loses only 0.03 in R@0.25 vs. 0.09 with COCO pretraining. These results underscore again the importance of leveraging FFDM-pretrained weights as much as possible, since they provide a stronger inductive bias for DBT and yield substantial gains in both low-data and low-annotation regimes.



\begin{table}[t]
\footnotesize
\setlength{\tabcolsep}{2pt}
\centering
\caption{Effect of z-resolution on M\&M-3D. \textbf{Mem.}: training GPU memory required. \textbf{Time}:  wall clock training time.}
\label{tab:ablation_num_slabs}
\begin{tabular}{cccccc}
\toprule
\textbf{z-Res.} & \textbf{Mem.} & \textbf{Time} & \textbf{R@0.25} & \textbf{R@0.5} & \textbf{AUC} \\ 
& \textbf{(GB)} & \textbf{(h)} & & & \\ 
\midrule
8  & 42 & 11& 0.69$_{\pm 0.022}$ & 0.76$_{\pm 0.020}$ & 0.94$_{\pm 0.005}$ \\
16 &84 &12 &0.77$_{\pm 0.021}$ & 0.82$_{\pm 0.019}$ & 0.95$_{\pm 0.004}$ \\
32 & 168 & 12 & 0.79$_{\pm 0.021}$ & 0.83$_{\pm 0.019}$ & 0.95$_{\pm 0.004}$ \\
\bottomrule
\end{tabular}
\end{table}

\begin{table}[t]
\footnotesize
\centering
\caption{Effect of fusion methods on M\&M-3D.}
\label{tab:ablation_compress}
\begin{tabular}{lccc}
\toprule
\textbf{Method} & \textbf{R@0.25} & \textbf{R@0.5} & \textbf{AUC} \\
\midrule
MeanPool  & 0.72$_{\pm 0.021}$ & 0.76$_{\pm 0.020}$ & 0.95$_{\pm 0.004}$ \\
MaxPool   & 0.77$_{\pm 0.021}$ & 0.81$_{\pm 0.021}$ & 0.95$_{\pm 0.005}$ \\
Wgt. Avg. & 0.77$_{\pm 0.021}$ & 0.82$_{\pm 0.019}$ & 0.95$_{\pm 0.004}$ \\
\bottomrule
\end{tabular}
\end{table}

\subsection{Ablation Studies}
\label{sec:ablation_studies}
\paragraph{z-Resolution.} \autoref{tab:ablation_num_slabs} examines the impact of downsampling z-resolution to 8, 16, or 32 slices. 
Higher resolution boosts recall and AUC, reinforcing the value of richer 3D information. However, increasing z-resolution also affects computational cost. If training resources pose a constraint,  M\&M-3D with 8 slices requires 42 GB GPU memory---only modestly higher than projection and slice-by-slice methods (30 GB)---while maintaining similar training time of 11 hours and significantly improving both classification and detection accuracy (0.94 AUC versus 0.92 AUC, 0.76 R@0.5 versus 0.74 R@0.5). Overall, we choose a z-resolution of 16 slices as it provides the best balance between performance and computational efficiency.

\paragraph{Fusion method.} \autoref{tab:ablation_compress} compares parameter-free strategies for fusing 3D features. 
MeanPool significantly harms detection, suggesting that equally weighting all slices causes non-suspicious slices to dilutes the representation with non-discriminative signal. MaxPool performs on-par with Wgt. Avg., suggesting that the representation of the most suspicious slice is the most important for malignancy determination. Notably, detection improves significantly, while AUC remains stable, suggesting that emphasizing suspicious slices in the 3D representation benefits detection more than classification.

\section{Conclusion}

We introduced M\&M-3D, a data-efficient detector for DBT that enables learned 3D reasoning without additional learnable parameters. M\&M-3D forms 3D representations through malignancy-weighted feature fusion and facilitates information mixing through the cascading heads. This design leverages 3D information while allowing direct transfer of FFDM weights. Our experiments highlight the benefits of learnable 3D reasoning (\autoref{sec:quantitative_performance}) and show that thoughtful model design can unlock its potential without compromising data efficiency (\autoref{sec:data_efficiency}).

Looking ahead, M\&M-3D paves the way for unified 2D-3D learning as the model design allows handling both DBT and FFDM without architectural changes. This flexibility enhances generalizability by enabling training on larger combined 2D-3D datasets. The unified representation also supports creation of models that can seamlessly reason across exams from different years despite heterogeneous modalities, mirroring the way radiologists leverage prior exams \citep{hayward2016improving}.


\bibliography{jmlr-sample}

\clearpage
\appendix
\section{Comparison to Literature on BCS-DBT}\label{apd:duke}
\autoref{tab:auc_comparison} demonstrates that M\&M-3D achieves state-of-the-art (SOTA) classification performance on BCS-DBT, with an AUC of 0.97 after finetuning on our  internal dataset. Remarkably, even without any DBT-specific training data, the model attains an AUC of 0.91, which surpasses several methods from recent years, including \cite{cantone2025deep} and \cite{kassis2024detection}. This finding highlights the strong transferability of representations learned from FFDM and their robustness to domain shift. With just 10\% of DBT training data, M\&M-3D reaches an AUC of 0.93, placing it on par with the top 3 methods on BCS-DBT from the DBTex challenge \citep{konz2023competition}: NYU BTeam, Zedus and Vicorob. Note that all 3 DBTex methods utilize internal data and then fine-tune on BCS-DBT train set. Meanwhile, M\&M-3D is only trained using our internal data. Our results suggest that M\&M-3D offers a scalable path to high-performance DBT classification, even in low-data regimes.
\begin{table}[h]
\centering
\caption{Classification performance on BCS-DBT. M\&M-3D performs comparably to top methods even with 0\% DBT data used. After finetuning with 100\% data, M\&M-3D achieves the highest AUC reported to date.(*) Methods not evaluated on the official test set. ($\dagger$)  As reported in \cite{terrassin2024thick}.}
\begin{tabular}{lc}
\hline
\textbf{Method} & \textbf{AUC} \\
\hline
\cite{tardy2021trainable}$^{*}$ & 0.73 \\
\cite{cantone2025deep} 3D-Vol. & 0.81 \\
\cite{cantone2025deep} 2D-Proj. & 0.84 \\
\cite{kassis2024detection} & 0.66 \\
\cite{park2023efficient}$^{*}$ & 0.85 \\
\cite{terrassin2024thick} & 0.90 \\
Zedus \citep{konz2023competition}$^{\dagger}$ & 0.92 \\
NYU BTeam \citep{konz2023competition}$^{\dagger}$ & 0.93 \\
Vicorob \citep{konz2023competition}$^{\dagger}$ & 0.93 \\
\cite{du2024sift} & 0.93 \\
\hline
\textbf{M\&M-3D}-0\% data & 0.91 \\
\textbf{M\&M-3D}-10\% data & 0.93 \\
\textbf{M\&M-3D}-100\% data & \textbf{0.97} \\
\hline
\end{tabular}
\label{tab:auc_comparison}
\end{table}

\autoref{tab:recall_comparison} reports malignant lesion recall, with false positive computed over all non-benign volumes. We remove benign cases from metric computation as DBTex challenge considers benign findings as true positive, whereas our setup considers these findings as hard false positives. For this comparison, we obtained \href{https://www.cancerimagingarchive.net/wp-content/uploads/team_predictions_bothphases.zip}{released predictions} from the top-5 DBTex methods and computed R@0.25 and R@0.5. The results show that M\&M-3D consistently outperforms top methods from the DBTex competition \citep{konz2023competition}. Direct transfer from FFDM (0\% DBT data) already achieves recalls superior to the 4th and 5th ranking teams (Prarit and UCLA-MII). With 10\% finetuning data, M\&M-3D surpasses all methods in R@0.5. When trained with the full DBT dataset, M\&M-3D achieves the best performance across both R@0.25 and R@0.50 metrics. 

Our results highlight that M\&M-3D provides consistent improvements across both classification and detection tasks, setting a new benchmark for cancer detection in DBT.

\begin{table}
\centering
\caption{Detection performance on BCS-DBT. We compare with top-5 methods from the DBTex Competition \cite{konz2023competition}. Direct transfer from FFDM using M\&M-3D outperforms the 4th and 5th ranking methods. With 10\% finetuning data, M\&M-3D outperforms all methods for R@0.5. With 100\% finetuning data, M\&M-3D achieves best performance across all metrics.}
\begin{tabular}{lcc}
\hline
\textbf{Method} & \textbf{R@0.25} & \textbf{R@0.5} \\
\hline
1. NYU B-Team      & 0.70 & 0.74 \\
2. ZeDuS           & 0.70 & 0.76 \\
3. Vicorob         & 0.74 & 0.77 \\
4. Prarit          & 0.41 & 0.56 \\
5. UCLA-MII        & 0.41 & 0.50 \\
\hline
\textbf{M\&M-3D}-0\% data    & 0.52 & 0.56 \\
\textbf{M\&M-3D}-10\% data    & 0.67 & 0.79 \\
\textbf{M\&M-3D}-100\% data  & \textbf{0.83} & \textbf{0.85} \\
\hline
\end{tabular}
\label{tab:recall_comparison}
\end{table}


\section{Detailed Performance Results Across Data and Annotation Regimes}
\label{apd:fig5}
We report the performance results used to generate \autoref{fig:3d_comparison}. In particular, \autoref{tab:data_det} and \autoref{tab:data_cls} report detection and classification performance, respectively, when varying data sizes from 10\% to 100\%. \autoref{tab:ann_det} reports detection performance when varying annotation sizes from 2\% to 40\% while keeping dataset size fixed at 100\%.
\begin{table}[ht]
\centering
\caption{Detection performance in terms of R@0.25 across different data regimes.}
\label{tab:data_det}
\begin{tabular}{lccc}
\toprule
\textbf{Method} & \textbf{10\%}& \textbf{40\%} & \textbf{100\%}\\
& \textbf{data} & \textbf{data} & \textbf{data} \\
\midrule
TimeSform    & 0.51 & 0.65 & 0.74 \\
QuerySummary & 0.51 & 0.70 & 0.77 \\
MLP-Regress  & 0.41 & 0.63 & 0.69 \\
\textbf{M\&M-3D}     & \textbf{0.61} &\textbf{0.71}& \textbf{0.77} \\
\bottomrule
\end{tabular}
\end{table}

\begin{table}[ht]
\centering
\caption{Classification performance in terms of breast AUC across different data regimes.}
\label{tab:data_cls}
\begin{tabular}{lccc}
\toprule
\textbf{Method} & \textbf{10\%}& \textbf{40\%} & \textbf{100\%}\\
& \textbf{data} & \textbf{data} & \textbf{data} \\
\midrule
TimeSform    & 0.90 & 0.93 & 0.95 \\
QuerySummary & 0.88 & 0.94 & 0.95 \\
MLP-Regress  & 0.83 & 0.94 & 0.95 \\
\textbf{M\&M-3D}      & \textbf{0.91} & \textbf{0.94} & \textbf{0.95} \\
\bottomrule
\end{tabular}
\end{table}

\begin{table}[ht]
\centering
\caption{Detection performance in terms of R@0.25 across different annotation (ann.) regimes.}
\label{tab:ann_det}
\begin{tabular}{lccc}
\toprule
\textbf{Method} & \textbf{2\%} & \textbf{16\%} & \textbf{40\%} \\
& \textbf{ann.} & \textbf{ann.} & \textbf{ann.} \\
\midrule
TimeSform    & 0.69 & 0.73 & 0.74 \\
QuerySummary & 0.64 & 0.74 & 0.77 \\
MLP-Regress  & 0.44 & 0.63 & 0.69 \\
\textbf{M\&M-3D}      & \textbf{0.74} & \textbf{0.77} & \textbf{0.77} \\
\bottomrule
\end{tabular}
\end{table}

\section{Usability Study on z-axis Localization}\label{apd:usability}
To assess the clinical sufficiency of single-slice z-axis localization, we conducted semi-structured interviews with five board-certified radiologists. Participants were asked about their decision-making process when reviewing DBT exams and their preferences for how AI-assisted findings should be presented. Through this usability study, we found that:
\begin{itemize}
    \item 5/5 radiologists emphasized that callback and diagnostic decisions are made primarily on the slice where the finding is most clearly visible. 
    \item 5/5 radiologists prefered the idea of indicating each finding with the most representative slice on the 3D volume, which allows them to quickly find the finding in the volume and make their diagnostic determination. 
    \item 2/5 radiologists reported using z-extent mainly to verify whether a finding is ``real'' or tissue overlap. 
    \item 5/5 radiologists agreed that highlighting the most suspicious slice would substantially reduce navigation time without compromising diagnostic confidence.
\end{itemize}

These results support our design choice of supervising z-axis localization on a single representative slice for each finding, showing that our formulation is consistent with established radiologist workflow.

\section{M\&M-3D Algorithm}\label{apd:code}
\autoref{lst:mm3d} highlights the key modifications introduced in M\&M-3D relative to M\&M for handling DBT data. For simplicity, we illustrate the pseudocode for a single DBT view. The same procedure is applied symmetrically across both CC and MLO views during training and inference.
\begin{lstlisting}[style=mypython, caption={Pseudo code showing key modifications of M\&M-3D on top of M\&M to maximize transfer learning from FFDM to DBT.}, label={lst:mm3d}]
class MM3DModel:
    def __init__(self, mm_2d):
        """init with mm_2d, which bundles:
          - proposal initializer -> (b0: Nx4, h0: NxD)
          - SelfAttn, CrossAttn # Eqs. (1), (3)
          - DynamicConv # Eq. (5)
          - ClsModule, RegModule # box cls. & reg. heads
          - MIL pooling (noisy-or, mean across CC/MLO)
          - get_alt_view_feats # retrieved feature from the other ipsilateral view
        No new learnable params are introduced.
        """
        self.mm_2d = mm_2d 

    def forward_one_view(self, dbt_image, cross_view_image):
        """simplify forward pass on one image
        """
        b = self.mm_2d.init_boxes
        h = self.mm_2d.init_features
        backbone_feat = self.mm_2d.backbone(dbt_image)

        w_last, z_last = None, None

        for i in range(self.mm_2d.num_heads):
            # Global & multi-view mixing
            h_prime = self.mm_2d.SelfAttn(h)    # Eq. (1)
            alt_view_feats = self.mm_2d.get_alt_view_feats(cross_view_image)
            h_tilde = self.mm_2d.CrossAttn(h_prime, alt_view_feats) # Eq. (2)

            # Slice-level Feature Interaction
            slice_h, slice_scores = [], []
            for s in range(backbone_feat.shape[1]):
                f_b_s = self.mm_2d.RoIAlign(backbone_feat, b)
                # mixing of 3D feature h_tilde and slice features f_b_s
                h_i_s = self.mm_2d.DynamicConv(h_tilde, f_b_s)     # Eq. (5)
                slice_h.append(h_i_s)
                m_i_s = self.mm_2d.ClsModule(m_i_s)
                slice_scores.append(m_i_s)

            # Slice-to-Volume Feature Fusion
            w = torch.softmax(torch.tensor(scores_slice), dim=0)
            # fusing slice_h to get a refined 3D representation
            h = (w * torch.tensor(slice_h)).sum(dim=0)      # Eq. (6)

            # Box and score refinement
            b = self.mm_2d.RegModule(h, b)
            m = self.mm_2d.ClsModule(h)

            # z-axis localization
            z = torch.argmax(w, dim=0)
            w_last, z_last = w, z

        # MIL aggregation (noisy-or for volume)
        image_score   = self.mm_2d.image_noisy_or(m)

        return {
            "boxes_2d": b,
            "boxes_score": m,
            "boxes_z": z_last,
            "slice_weights": w_last,
            "image_score": image_score,
        }

    def loss(self, outputs, targets):
        """modified loss with z-localization supervision
        """
        L_mm = self.mm_2d.loss(outputs, targets)
        if targets.has_slice_annotation():
            L_z = F.cross_entropy(outputs["slice_weights"], targets.gt_slice)  # Eq. (7)
            return L_mm + L_z
        return L_mm
\end{lstlisting}

\end{document}